\documentclass{midl} 

\usepackage{makecell}
\usepackage{mwe} 
 \usepackage{algorithm}
\usepackage{algpseudocode}
\usepackage{caption}
\usepackage{float} 
\usepackage{booktabs}
\usepackage[colorinlistoftodos]{todonotes}
\usepackage{graphicx}   
\usepackage{algorithm}  
\usepackage{algorithmicx}
\usepackage{algpseudocode} 
\usepackage{amsmath}  
\makeatletter
\expandafter\let\csname enumerate*\endcsname\relax
\expandafter\let\csname endenumerate*\endcsname\relax
\makeatother

\usepackage[inline]{enumitem}
\usepackage{soul}
\sethlcolor{cyan}
\usepackage{multirow}
\usepackage[most]{tcolorbox}

\jmlrproceedings{}{}

\jmlrpages{}

\makeatletter
\renewcommand*{\@titlefoot}{}
\makeatother

\title[RadVLM-GRPO]{Enhancing Radiology Report Generation and Visual Grounding using Reinforcement Learning}

\midlauthor{\Name{Benjamin Gundersen\nametag{$^{1}$}\midljointauthortext{Contributed equally.}}  \Email{benjamin.gundersen@uzh.ch}\\
\Name{Nicolas Deperrois\nametag{$^{1}$}}\midlotherjointauthor \Email{nicolas.deperrois@uzh.ch}\\
\Name{Samuel Ruiperez-Campillo}\nametag{$^{2}$} \Email{samuel.ruiperezcampillo@inf.ethz.ch}\\
\Name{Thomas M. Sutter}\nametag{$^{2}$} \Email{thomas.sutter@inf.ethz.ch}\\
\Name{Julia E. Vogt}\nametag{$^{2}$}\Email{julia.vogt@inf.ethz.ch}\\
\Name{Michael Moor}\nametag{$^{3}$}\Email{michael.moor@bsse.ethz.ch}\\
\setsharedfootnotesymbol
\Name{Farhad Nooralahzadeh\nametag{$^{1,4}$}}\midljointsharedauthortext{Shared senior authorship.}\Email{farhad.nooralahzadeh@uzh.ch}\\
\Name{Michael Krauthammer\nametag{$^{1}$}\midlothersharedauthor}\Email{michael.krauthammer@uzh.ch}\\
\addr $^{1}$ Department of Quantitative Biomedicine, University of Zurich, Zurich, Switzerland \\
\addr $^{2}$ Department of Computer Science, ETH Zurich, Zurich, Switzerland \\
\addr $^{3}$ Department of Biosystems Science and Engineering, ETH Zurich, Basel, Switzerland\\
\addr $^{4}$ Institute of Computer Science, Zurich University of Applied Sciences, Zurich, Switzerland
}

\begin{document}

\maketitle

\begin{abstract}
Recent advances in vision–language models (VLMs) have improved Chest X-ray (CXR) interpretation in multiple aspects. However, many medical VLMs rely solely on supervised fine-tuning (SFT), which optimizes next-token prediction without evaluating answer quality. In contrast, reinforcement learning (RL) can incorporate task-specific feedback, and its combination with explicit intermediate reasoning (``thinking'') has demonstrated substantial gains on verifiable math and coding tasks. To investigate the effects of RL and thinking in a CXR VLM, we perform large-scale SFT on CXR data to build an updated RadVLM based on Qwen3-VL, followed by a cold-start SFT stage that equips the model with basic thinking ability. We then apply Group Relative Policy Optimization (GRPO) with clinically grounded, task-specific rewards for report generation and visual grounding, and run matched RL experiments on both domain-specific and general-domain Qwen3-VL variants, with and without thinking. Across these settings, we find that while strong SFT remains crucial for high base performance, RL provides additional gains on both tasks, whereas explicit thinking does not appear to further improve results. Under a unified evaluation pipeline, the RL-optimized RadVLM models outperform their baseline counterparts and reach state-of-the-art performance on both report generation and grounding, highlighting clinically aligned RL as a powerful complement to SFT for medical VLMs\footnote{Code is available at \url{https://github.com/uzh-dqbm-cmi/RadVLM-GRPO} and the updated SFT and RL models will be released under a new version at  \url{https://physionet.org/content/radvlm-model}}.
\end{abstract}


\begin{figure}[h]
\floatconts
  {fig:presentation-radvlm}
  {\caption{ \textbf{Optimization of RadVLM with GRPO.}
  We optimize RadVLM with GRPO by sampling $N$ outputs from the model conditioned on an image-instruction pair. We then compare the outputs to the ground truth with a task-specific reward: RadCliQ for report generation and IoU-based metric (soft-F1) for visual grounding. These rewards provide an update signal to optimize RadVLM.
    }}  {\includegraphics[width=1.0\linewidth]{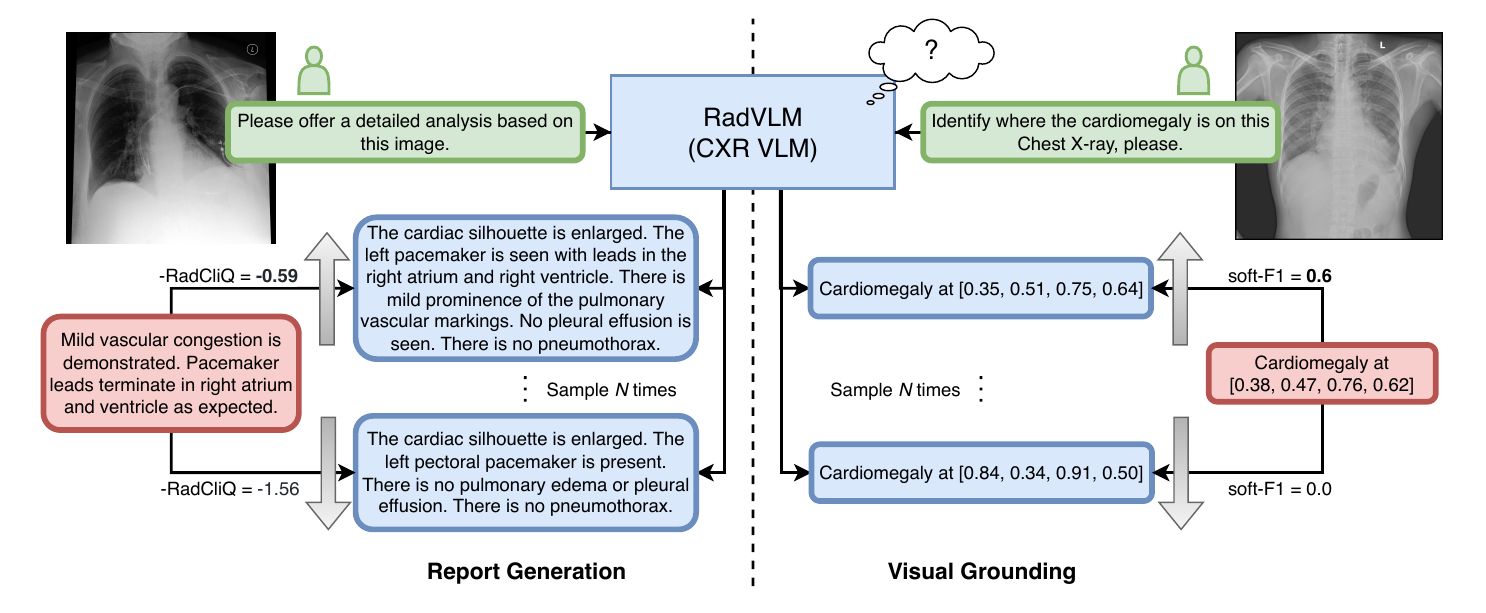}}
\end{figure}

\section{Introduction}
Reinforcement learning has emerged as a central paradigm in modern reasoning-focused foundation models~\cite{guo2025deepseek, zheng2025learning}. Recent work in large language models (LLMs) has shown that RL, combined with explicit intermediate reasoning ("thinking"), can substantially improve performance on math and coding tasks~\cite{shao2024deepseekmath, guo2025deepseek, yang2025qwen3, yu2025dapo}. Rather than solely relying on supervised fine-tuning (SFT), these methods employ RL algorithms such as Group Relative Policy Optimization (GRPO)~\cite{shao2024deepseekmath} to reward outputs. 
However, current medical VLMs specialized in CXR interpretation are typically optimized via image-instruction SFT-only~\cite{chen2024chexagent, sellergren2025medgemma, deperrois2025radvlm}. Recent work applied GRPO in the medical and CXR domain on closed-ended tasks, such as visual question answering (VQA) and classification with predefined labels~\cite{lai2025med, pan2025medvlm, fan2025chestx}.

In contrast, the workflow of radiologists centers on producing open-ended, free-text reports, which could in principle be supported by VLMs trained directly on report generation~\cite{tanno2025collaboration}. However, the SFT objective -- next-token prediction -- is agnostic to the clinical utility of the generated report: it optimizes local likelihood rather than sequence-level, clinically meaningful criteria that RL could potentially optimize. For visual grounding, which plays a crucial role for radiology assistance~\cite{lee2022localization}, a similar limitation arises:  token-level SFT-only matches coordinate strings, while the true objective is a geometric overlap between predicted and reference bounding boxes. These gaps motivate moving beyond SFT-only optimization and systematically studying post-training RL for report generation and visual grounding. 

In that regard, previous work has explored RL for CXR report generation through offline direct preference optimization (DPO)~\citep{hein2024chexalign, liang2025chexpo}, and more recently through online GRPO~\citep{lin2025foundation}. Yet, the benefits of task-aligned RL for open-ended reporting and grounding remain insufficiently understood. In particular, it is unclear (i) how to best design rewards that jointly reflect lexical similarity and clinical correctness for reports, and continuous overlap-based feedback for grounding; (ii) whether performance gains arise from the RL objective alone or from its combination with preceding thinking; and (iii) to what extent SFT on domain-specific data is necessary when performing RL on that same domain.

To address these questions, we build an updated version of the state-of-the-art multitask conversational CXR-VLM RadVLM~\citep{deperrois2025radvlm}, based on Qwen3-VL~\citep{bai2025qwen3vltechnicalreport} and focus on its GRPO-based optimization for report generation and visual grounding, as visualized in \figureref{fig:presentation-radvlm}. For report generation, we employ a clinically grounded reward, RadCliQ~\citep{yu2023evaluating}, while for visual grounding, we design a continuous IoU-based reward. We further synthesize thinking data, thereby equipping RadVLM with explicit thinking and non-thinking variants, and optimize both with GRPO. In addition, we apply the same RL procedure to out-of-domain Qwen3-VL-8B Instruct and Thinking models to assess how what extent a general-domain VLM can be adapted to the CXR domain by using RL alone. Under a unified evaluation pipeline, we find that training RadVLM with GRPO for grounding and report generation consistently improves all metrics over its SFT counterpart and outperforms state-of-the-art baselines on most metrics, while explicit thinking yields at best marginal additional gains over direct-answer variants.

\section{Related Work}

\subsection{RL and thinking for LLMs}
Recent advances in LLM research show that RL can markedly improve performance, especially when combined with explicit thinking and verifiable rewards. The OpenAI thinking series~\citep{openai2024learning}, open-weight models such as DeepSeek-R1~\citep{liu2024deepseek, guo2025deepseek}, and Qwen3~\citep{yang2025qwen3} typically combine a cold-start SFT stage, which teaches the model how to explicitly reason before providing an answer, followed by online RL. These RL-with-verifiable-reward (RLVR) pipelines yield large gains on tasks such as math and code, particularly when encouraging chain-of-thought reasoning before providing the final answers~\citep{guo2025deepseek, muennighoff2025s1, hu2025open}. However, their success strongly depends on reliable reward signals; for open-ended tasks, reward design is more challenging, and improvements are less consistent between tasks~\cite{cheng2025revisiting}.

\subsection{RL and thinking for VLMs}
Inspired by text-based results, the combination of RL and thinking has also been explored for VLMs, with a strong emphasis on visual-math tasks. Vision-R1~\citep{huang2025vision} synthesizes multimodal chain-of-thought data and applies GRPO to Qwen2.5-VL~\citep{bai2025qwen2}, while VL-Rethinker~\citep{wang2025vl} augments training with replay and "rethinking" triggers to improve visual-math. Beyond math, online RL also helps perception and grounding: Visual-RFT~\citep{liu2025visual} and VLM-R1~\citep{shen2025vlm} report improved localization and visual understanding over SFT with limited additional data. Still, it remains unclear how much benefit currently comes from explicit thinking versus reward shaping and exploration across different domains. Experiments on VLMs of similar size as RadVLM indicate that, for some tasks, applying GRPO to models with explicit chain-of-thought can even underperform models that directly produce the final answer~\cite{li2025think, lai2025med}.
Moreover, Qwen3-VL~\citep{bai2025qwen3vltechnicalreport} is obtained through multi-stage training (including RL), and its instruct (non-thinking) variant outperforms the thinking variant on some tasks, including Document Understanding, 2D / 3D Grounding, Perception with Tools, and Multi-Modal Coding (see Appendix \tableref{tab:qwen-q3vl-summary}).

\subsection{VLMs for radiology}
Many CXR VLMs are trained via supervised instruction tuning on multi-task radiology data. General medical models such as LLaVA-Med~\citep{li2023llava-med}, MedGemma~\citep{sellergren2025medgemma}, as well as CXR-focused systems such as CheXagent~\citep{chen2024chexagent}, MAIRA-2~\citep{Bannur2024-ek}, and RadVLM~\citep{deperrois2025radvlm}, rely on a large instruction corpora covering report generation, classification, and grounding~\citep{deperrois2025radvlm-dataset}.

Med-R1~\citep{lai2025med}, MedVLM-R1~\citep{pan2025medvlm}, and ChestX-Reasoner~\citep{fan2025chestx} apply GRPO-style training to close-ended CXR/medical VQA, demonstrating gains in diagnostic accuracy and structured reasoning. For open-ended report generation, RL has so far predominantly been used in the form of offline direct preference optimization (DPO): CheXalign~\citep{hein2024chexalign} and CheXPO~\citep{liang2025chexpo} automatically construct preference pairs but do not study online RL or explicit thinking. DeepMedix-R1~\citep{lin2025foundation} is the closest to our setting, applying online GRPO with thinking to report generation; however, it optimizes a combination of lexical rewards without incorporating an explicit clinical signal, and does not disentangle gains due to GRPO versus explicit chain-of-thought.

\section{Methods}
We build upon standard components used in modern VLMs and combine SFT and GRPO-based RL~\cite{shao2024deepseekmath, guo2025deepseek}. A brief summary is provided in Appendix~\ref{appendixsec:preliminaries}.

\subsection{Training data}
\label{sec:data}
We use two types of datasets: an instruction dataset and a cold-start dataset. The instruction dataset is the RadVLM dataset~\citet{deperrois2025radvlm}, a multimodal corpus of over one million image-text pairs. Each sample consists of a single frontal CXR image paired with a user-assistant interaction. The dataset spans four complementary instruction types: (i) free-text report generation, where the assistant produces the findings section from a single image; (ii) abnormality classification; (iii) visual grounding with bounding-box supervision, covering anatomical, abnormality, and phrase-level localization; and (iv) multi-turn conversations generated with GPT-4o. These components jointly support the training of models with fine-grained visual understanding and robust conversational behavior in radiology. Full details on the dataset construction and preprocessing are provided in~\citet{deperrois2025radvlm-dataset, deperrois2025radvlm}.

In addition, we create a cold-start dataset consisting of thinking examples. We prompt Qwen3-VL-235B-Instruct~\citep{bai2025qwen3vltechnicalreport} to generate a thinking trajectory given the image and ground truth for both report generation and grounding (see Appendix \ref{appendix-sec:cold-start-prompts} for prompt details). We then separate the thinking part from the final answer using \texttt{</think>} and collect a total of around 28k such data points.

\begin{figure}[t]
 \floatconts
  {fig:model-overview}
  {\caption{\textbf{Overview of Model Checkpoints and Training Pipeline.} The diagram illustrates how different model variants are initialized and refined with SFT and RL.}}
  {\includegraphics[width=0.9\linewidth]{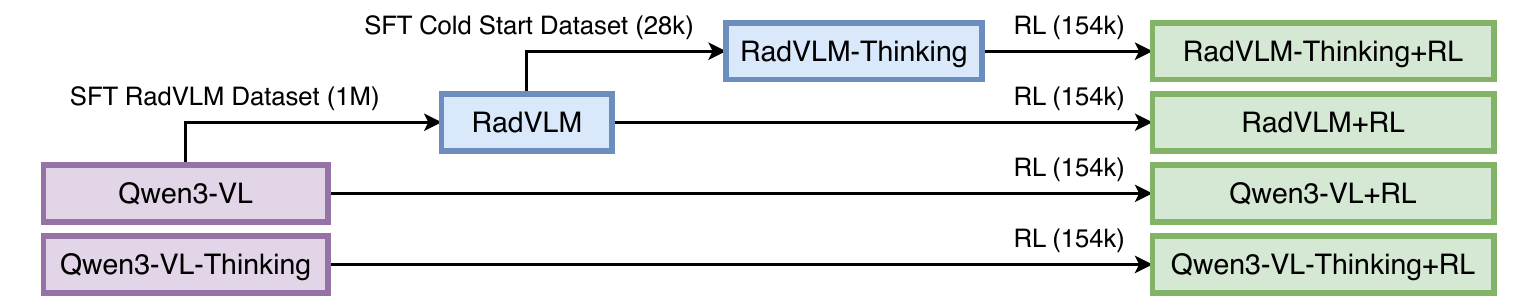} }
\end{figure}

\subsection{Model Training}
We train multiple models in stages. We first create SFT checkpoints (RadVLM and RadVLM-Thinking) and then enhance these checkpoints with RL. \figureref{fig:model-overview} summarizes the SFT and RL pipeline. Throughout, Qwen3-VL refers to the 8B Instruct model.

We begin by performing SFT on the RadVLM dataset (see Section~\ref {sec:data}). Starting from Qwen3-VL-8B-Instruct, we finetune on the question-answer pairs, yielding an updated version of RadVLM, as visualized in \figureref{fig:model-overview} (see Appendix~\ref{tab:radvlm-comparison-visual-grounding} for comparison to the original RadVLM model). Based on this SFT checkpoint, we perform a second SFT stage, where we combine datapoints from the cold-start dataset with a small subset of direct question-answer datapoints, resulting in RadVLM-Thinking.  The cold-start datapoints have a \texttt{/think} suffix to differentiate them.
The data mix between cold-start and SFT datapoints in that phase is designed to prevent potential catastrophic forgetting. 

Following the SFT phase, we continue with RL to further optimize our model and align it with radiology-specific objectives.
We perform GRPO training on continuous, task-specific rewards.
For all models, the reward is computed on the final answer. For thinking models, the final answer is defined as the text following the \texttt{</think>} delimiter; if no such final answer is produced, a reward of $0$ is assigned, unless noted otherwise. For non-thinking models, the entire model output is treated as the final answer.


Given our multitask setting (report generation and visual grounding), we design rewards aligned with both tasks. For report generation, we use as a reward RadCliQ~\citep{yu2023evaluating}, a composite metric consisting of BERTScore~\citep{zhang2020BERTScore}, CheXbert~\citep{smit2020chexbert} vector similarity, and RadGraph-F1~\citep{jain2021radgraph, delbrouck2024radgraph}, which has been shown to align better with experts than any of these alone~    \citep{yu2023evaluating}.
For visual grounding, we use a \textit{Hungarian-matched soft-F1} reward that provides smooth, per-example feedback for bounding-box alignment. For each image, we compute pairwise IoUs between predicted and reference boxes and obtain a one-to-one assignment via the Hungarian algorithm. Matched predictions contribute proportionally to their IoU (soft true positives), while unmatched predictions and ground-truth boxes act as false positives and false negatives. In contrast to mAP, which counts a prediction as correct only if its IoU exceeds a fixed threshold, this reward gives partial credit for close matches and avoids discontinuities, offering a more stable and informative signal for RL. Our design follows DETR's matching strategy~\citep{carion2020end} and draws inspiration from IoU-weighted metrics such as Panoptic Quality~\citep{kirillov2019panoptic}.

\section{Experiments}

\subsection{Training settings}

We start with Qwen3-VL-8B-Instruct~\citep{bai2025qwen3vltechnicalreport} and perform SFT on the entire RadVLM instruction dataset for two epochs with a learning rate of $1e-5$, train batch size of $8*64$ ($64$ GPUs), using LLaMA-Factory~\cite{zheng2024llamafactory}, resulting in an updated RadVLM. Next we add thinking capabilities to RadVLM by performing another round of SFT with the same settings on a mix of the cold-start dataset and RadVLM instruct dataset to prevent catastrophic forgetting.

Finally, we perform GRPO individually for each task and model variant, with 300 steps and a batch size of 512 prompts, processing a total of 154k processed prompt-image pairs. We create 8 rollouts per prompt-image pair, the KL penalty coefficient is 0.01, and we use asymmetric clipping~\cite{yu2025dapo}. Please refer to Appendix \tableref{suptab:verl-parameters} for a detailed overview of hyperparameters. To apply GRPO~\cite{shao2024deepseekmath} we use the verl library~\citep{sheng2024hybridflow}. As rewards, we use RadCliQ for report generation and soft-F1 for grounding.

We use the RadCliQ implementation provided in RadEval~\citep{xu2025radevalframeworkradiologytext}, with RadGraph-XL~\citep{delbrouck2024radgraph} as our chosen RadGraph model. Since lower is better for RadCliQ, we use -RadCliQ as a reward. Note that $-$RadCliQ frequently computes values lower than $0$ and thus we provide a reward of $-3$ if there is no final answer. Since RadGraph does not support batch inference, we spin up 32 individual copies of RadCliQ such that the reward computation does not become a bottleneck.

\subsection{Baseline models }
We evaluate a broad set of baseline models, including Qwen3-VL variants~\cite{yang2025qwen3}, CheXagent-2~\cite{chen2024chexagent}, MedGemma~\cite{sellergren2025medgemma}, DeepMedix-R1~\cite{lin2025foundation}, MAIRA-2~\cite{Bannur2024-ek}, and Llava-Rad~\cite{zambrano2025clinically}. We denote our GRPO-trained variants with ``+RL'', and ``Thinking'' models generate intermediate reasoning before providing the final answer. The prompts and Hugging Face (HF) links are provided in Appendix~\ref{appendix-sec:model-prompts} and~\ref{appendix-sec:hf-links} respectively.

\subsection{Evaluation}
\label{subsec:evaluation}
We evaluate on the report generation test set of RadVLM, derived from MIMIC-CXR \cite{deperrois2025radvlm-dataset}. The test set consists of single frontal chest X-rays paired with findings, with references to prior studies removed. We use various NLP and clinical metrics to evaluate report generation (see Appendix~\ref{sec:evaluation-metrics-details} for details). To quantify textual similarity, we report BERTScore~\citep{zhang2020BERTScore} and ROUGE-L~\citep{lin2004rouge}. To evaluate clinical correctness, we report RadGraph F1~\citep{jain2021radgraph} using RadGraph-XL~\citep{delbrouck2024radgraph}, CheXbert F1~\citep{smit2020chexbert}, RadCliQ~\citep{yu2023evaluating} and GREEN~\citep{ostmeier2024green}.

For the grounding task, we report mAP@0.5, computed by considering a prediction correct if its IoU with an unmatched reference box exceeds~0.5, in line with previous work on object detection benchmarks~\citep{everingham2010pascal,lin2014microsoft}. In particular, we report this metric on the test splits for anatomical grounding (Chest Imagenome dataset), abnormality grounding (VinDr-CXR dataset) and phrase grounding (MS-CXR and PadChest-GR datasets), following the same procedure as in \citet{deperrois2025radvlm}. For thinking models, we extract the final answer following \texttt{</think>} and evaluate it using the same procedure as for direct answers.

\subsection{Results}
\subsubsection{GRPO Consistently improves RadVLM and Qwen3-VL Performance}

For report generation (\tableref{tab:reportgeneration-results}), GRPO substantially improves performance for both RadVLM and Qwen3-VL models. Notably, the held-out metric GREEN improves, indicating robustness with respect to the reward signal. Among all evaluated models, GRPO-optimized RadVLM variants achieve the strongest overall performance. However, despite these gains, RadVLM models remain behind in terms of CheXbert-macro. 
For Qwen3-VL variants, applying GRPO alone leads to large improvements, even outperforming the SFT-only RadVLM variant in some metrics. Comparing thinking and non-thinking variants, thinking models generally perform slightly worse than their instruct counterparts, with the exception of RadVLM+RL in report generation where both variants achieve comparable performance.

For visual grounding (\tableref{tab:grounding-results}), general-domain models perform poorly on CXR grounding, with a small but consistent advantage for the non-thinking variant -- in line with Qwen3-VL's grounding performance reported in~\citet{bai2025qwen3vltechnicalreport}. After GRPO training, both variants improve across all tasks, again with the non-thinking model ahead, indicating that RL alone can strongly enhance out-of-domain grounding -- yet they still remain slightly below SFT-only RadVLM. Applying RL on top of RadVLM further increases mAP, showing that even a well-trained in-domain model benefits from task-aligned RL. In contrast, the RL-optimized thinking variant of RadVLM does not outperform its direct-answer counterpart.

\begin{table}[h]
\setlength\tabcolsep{2pt}

 \floatconts
 {tab:reportgeneration-results}%
 {\caption{\textbf{Report generation performance}. ROUGE-L (R-L), BERTScore (B-S), CheXbert micro and macro (CXb), RadGraph F1 (RGF1), GREEN (GRN), and RadCliQ (RCQ) are evaluated on the report generation test set of RadVLM, derived from MIMIC-CXR. For RadCliQ, lower is better ($\downarrow$); for all other metrics, higher is better. RL denotes training with GRPO. All models are evaluated using our pipeline. \textbf{Bold} values indicate the highest performance, while \underline{underlined} values represent the second-best performance. Evaluation prompts are listed in Appendix \ref{appendix-sec:model-prompts}.}}%
 {
 \begin{tabular}{@{}p{0.3\linewidth}
 p{0.06\linewidth}
 p{0.08\linewidth}
 p{0.08\linewidth}
 p{0.08\linewidth}
 p{0.08\linewidth}
 p{0.08\linewidth}
 p{0.08\linewidth}
 p{0.08\linewidth}@{}}
 \toprule
\bfseries Model & \bfseries Size (B) &\bfseries R-L & \bfseries B-S & \bfseries CXb-micro & \bfseries CXb-macro & \bfseries RGF1 & \bfseries GRN & \bfseries RCQ$\downarrow$ \\
\midrule
MedGemma-pt & 4 & 20.7 & 47.7 & 49.8 & 32.5 & 15.5 & 21.9 & 1.37 \\
MedGemma-it  & 27& 15.9 & 31.3 & 47.0 & 31.5 & 12.0 & 23.3 & 1.79 \\
MAIRA-2 &7 &17.7 & 46.6 & 52.1  & 35.8 & 12.9 & 21.3 & 1.42 \\
CheXagent-2 & 3 & 22.5 & 37.4 &  54.5 & \underline{38.7} & 20.1 & 29.9 & 1.45 \\
DeepMedix-R1 &7& 22.3 & 52.8 & 48.2 &  28.3 & 18.6 & 30.0 & 1.23\\
LLaVA-Rad &7& 22.2 & 48.9 & 53.3 &  \textbf{39.2} & 16.8 & 28.6 & 1.34 \\
\midrule
Qwen3-VL &8& 14.0 & 42.0 & 35.3  &  20.4 & 10.8 & 21.9 & 1.67 \\
Qwen3-VL-Thinking &8& 15.0 & 43.1 & 34.6 & 18.5   & 10.9 & 18.5 & 1.63 \\
Qwen3-VL+RL  &8& 24.7 & 55.8 & 47.4 & 22.3 & 22.5 & 25.8 & 1.05 \\
Qwen3-VL-Thinking+RL  &8& 23.3 & 55.0 & 44.4 & 22.4 & 21.0 & 24.7 & 1.12 \\
\midrule
RadVLM  &8& 26.0 & 53.3 & 49.0 & 30.5 & 19.0 & 29.1 & 1.14 \\
RadVLM+RL   &8& \underline{29.9} & \textbf{59.2} & \textbf{57.0} & 33.4 & \textbf{25.8} & \underline{32.7} & \textbf{0.86} \\
RadVLM-Thinking+RL  &8& \textbf{30.0} & \underline{59.0}  & \underline{56.3} & 33.6 & \underline{25.7} & \textbf{32.9} & \underline{0.87} \\
\bottomrule
\end{tabular}}
\end{table}

\begin{table}[t]
\floatconts
 {tab:grounding-results}%
 {\caption{\textbf{Visual grounding performance measured in mAP (\%)}. Mean average precision (mAP@0.5) scores for anatomical (Chest-Imagenome test dataset), abnormality (VinDr-CXR), and phrase grounding (Phrase$_{\text{MS}}$ for MS-CXR and Phrase$_{\text{Pad}}$ for PadChest-GR) tasks across various models. \textbf{Bold} values indicate the highest performance, while \underline{underlined} values represent the second-best performance. For Qwen3-VL, we specify in the prompt that the model should output boxes in the expected format (see Appendix \ref{appendix-sec:model-prompts}).}}%
{\begin{tabular}{p{0.35\linewidth}
 p{0.13\linewidth}
 p{0.13\linewidth}
 p{0.13\linewidth}
 p{0.13\linewidth}}
\toprule
\bfseries Model & \bfseries Anatomy & \bfseries Abnorm.  & \bfseries Phrase$_{\text{MS}}$  & \bfseries Phrase$_{\text{Pad}}$\\ 
\midrule
MAIRA-2 & 19.8 & 11.3 & 80.1 & 38.8 \\ 
\midrule
Qwen3-VL & 11.1 & 5.1 & 19.0 & 10.7 \\
Qwen3-VL-Thinking & 8.7 & 1.2 & 14.6 & 10.7 \\
Qwen3-VL+RL & 79.0 & 36.0 & 79.4 & 55.3 \\
Qwen3-VL-Thinking+RL & 68.8 & 27.6 & 76.1 & 42.0 \\ 
\midrule
RadVLM & 82.1 & \underline{44.2} & 84.6 & 59.0 \\ 
RadVLM+RL & \underline{84.5} & \textbf{45.9} & \textbf{87.9} & \textbf{63.0} \\ 
RadVLM-Thinking+RL & \textbf{84.9} & 43.9 & \underline{86.4} & \underline{60.5} \\
\bottomrule
\end{tabular}}%
\end{table}

\subsubsection{Reward choice influences report generation quality and length}

\begin{table}[t!]
\setlength\tabcolsep{2pt}

 \floatconts
 {tab:reportgeneration-rewards-comparison}%
 {\caption{\textbf{Effect of reward choice on report generation performance}. We compare RadVLM+RL with four different rewards, namely RadCliQ (RCQ), BERTScore (B-S), RadGraph-F1 (complete; RGF1-C), and GLEU across all evaluation metrics. The setting is the same as in \tableref{tab:reportgeneration-results}, with the addition of the average number of characters in the response (ANC). \textbf{Bold} values indicate the highest performance, while \underline{underlined} values represent the second-best performance, and $*$ indicates that the evaluated metric is used as a training reward.}}%
 {
 \begin{tabular}{@{}p{0.28\linewidth}
 p{0.08\linewidth}
 p{0.08\linewidth}
 p{0.08\linewidth}
 p{0.08\linewidth}
 p{0.08\linewidth}
 p{0.08\linewidth}
 p{0.08\linewidth}
 p{0.08\linewidth}@{}}
 \toprule
\bfseries Model  &\bfseries R-L & \bfseries B-S & \bfseries CXb-micro & \bfseries CXb-macro & \bfseries RGF1 & \bfseries GRN & \bfseries RCQ$\downarrow$ & ANC \\
\midrule
RadVLM  & 26.0 & 53.3 & 49.0 & 30.5 & 19.0 & 29.1 & 1.14 & 207\\
\midrule
+ RCQ & \textbf{29.9} & \underline{59.2} & \textbf{57.0} & 33.4 & \textbf{25.8} & 32.7 & \textbf{0.86}* & 171\\
+ B-S & 29.0 & \textbf{61.6}* & \underline{56.5} & \textbf{35.1} & 23.9 & \underline{33.1} & 0.93 & 206\\
+ RGF1-C & 22.8 & 48.8 & 34.1 & 18.4 & 20.1* & 23.5 & 1.20 & 93\\
+ GLEU & \underline{29.8} & \underline{59.2} & 56.0 & \textbf{35.1} & \underline{25.3} & \textbf{33.7} &  \underline{0.92} & 209    \\
\bottomrule
\end{tabular}}
\end{table}

We explore the impact of the reward function on the report generation task, comparing RadCliQ, BERTScore, RadGraph-F1, and GLEU~\cite{wu2016googles} in \tableref{tab:reportgeneration-rewards-comparison}. We evaluate the obtained RL-optimized RadVLM models on all report generation metrics. We find that RadCliQ, BERTScore, and GLEU rewards improve all metrics, with RadCliQ ranking highest in 4 out of the 7 metrics. However, training with RadGraph-F1 only improves the corresponding metric itself and leads to very short reports, revealing strong signs of reward hacking. Since GREEN is prone to length reward-hacking \cite{hein2024chexalign}, the higher GREEN scores of BERTScore and GLEU might be partially attributable to longer answers. Together, these results indicate that the choice of reward to optimize report generation via GRPO is crucial to avoid reward hacking and to improve performance in both lexical and clinical aspects. 

\subsubsection{Thinking Models Show Slower Reward Convergence}

We plot the evolution of the reward and response length during the RL training of RadVLM and Qwen3-VL models in \figureref{fig:rl_training}. We observe that the rewards of the Thinking variant (orange and red curves) remain slightly lower than the non-thinking versions, before eventually catching up. The RadVLM models (red and green) start out with a higher reward than Qwen3-VL models due to their initial in-domain SFT stage. For report generation, we notice that Qwen3-VL+RL (\figureref{fig:rl_training}c, blue curve) initially generates long reports and become shorter after a few RL steps, due to the RadCliQ optimization that discourages unwanted statements. Finally, thinking models tend to generate a relatively stable amount of tokens over RL steps (\figureref{fig:rl_training}c-d).

\begin{figure}[!t]
\floatconts
  {fig:rl_training}
  {\caption{
    \textbf{Training dynamics with GRPO.} Training rewards and response lengths are shown for report generation (a, c) and visual grounding (b, d) as a function of RL steps. 
    }}
  {\includegraphics[width=1.0\textwidth]{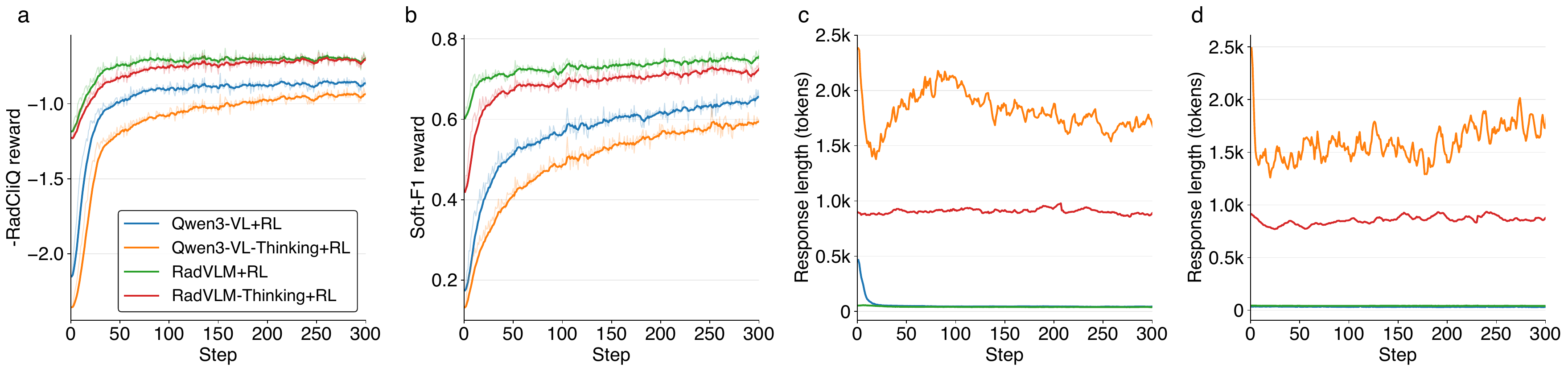}}
\end{figure}

\section{Conclusion, Discussion, Future Work}

In this work, we explored how RL can enhance the report generation and visual grounding capabilities of a foundation model for radiology. We show that by careful choice of rewards, the GRPO algorithm can further optimize RadVLM on tasks on which the SFT version was already performing well. This demonstrates that medical VLMs can be subjected to clinically relevant optimization enabled by modern RL techniques, overcoming limitations of SFT. Using the same optimization process, we also explored the potential benefits of thinking and found that in our settings, direct answers provide similar or better performance with lower inference cost. 

A potential limitation of our study is the choice of cold-start generation strategy. We chose to generate CoT by providing the prompt combined with the image and the expected ground truth, potentially biasing the model toward generating CoT that directly leads to that specific answer. However, we also explored other strategies that do not provide the ground truth and filter generated CoT-answer pairs based on the performance of the answer (see Appendix~\ref{appendix-sec:cold-start-prompts}). We observed that the thinking model underperforms compared to the non-thinking variant, consistent with Qwen3-VL on some tasks such as 2D/3D grounding \cite[\tableref{tab:qwen-q3vl-summary},][]{bai2025qwen3vltechnicalreport} and related work~\cite{li2025think, lai2025med}. Many factors could be at play here: initial performance degradation due to cold-start fine-tuning, the thinking traces could be non-informative for the tasks at hand, a lack of vision modality in the thinking process, or the pre-training and post-training processes of VLMs such as Qwen3-VL lack radiology data compared to math and code~\cite{cheng2025revisiting}. In this vein, future work could involve curating examples of thinking designed by expert radiologists and optimizing multi-turn conversations that include questions, feedback, and tools with which the model can manipulate images (cropping, zooming, drawing) and re-observe~\cite{zheng2025deepeyes, su2025openthinkimg, bai2025qwen3vltechnicalreport}. 

Another path for further exploration is LLM-as-Judge rewards, such as GREEN~\cite{ostmeier2024green}. LLM-as-Judge approaches require additional care to prevent reward hacking~\citep{hein2024chexalign, zhao2025one}, to design robust prompts, and to select appropriate models~\cite{li2025llms}. Yet, LLM-as-Judge is a powerful tool and could potentially enable optimization of multi-turn conversations by simultaneously serving as the reward mechanism and a conversational partner for the model under optimization.

Overall, by performing large-scale training on CXR data, our study highlights the importance of cooperation between fine-tuning and reinforcement learning in the development of medical VLMs. Although some questions remain open, we hope that our findings offer a solid foundation for future work on optimizing clinically aligned multimodal models.

\clearpage  
\midlacknowledgments{This work was supported as part of the Swiss AI Initiative by a grant from the Swiss National Supercomputing
Centre (CSCS) under project ID a135 on Alps. BG received support from the Swiss National Science Foundation (SNSF) grant 10003518. ND received support from RADICAL (Project-Call 2024.1, ID:9), funded by the DIZH.}

\bibliography{main}

\clearpage
\appendix

\section{Preliminaries}
\label{appendixsec:preliminaries}
\subsection{Vision-Language Model}
Vision-language models (VLMs) jointly model the visual and textual features in an autoregressive manner. Given an image $v$ and a text $t$, the visual input is encoded into a sequence of visual embeddings $x_v = (x_{v,1}, \ldots, x_{v,L_v})$ using a vision encoder and the text is tokenized, resulting in $L_t$ tokens $x_t = (x_{t,1}, \ldots, x_{t,L_t})$.

The model predicts the next textual token conditioned on all preceding text tokens
and the visual embeddings, defining the joint probability of the text sequence as

\begin{align}
p(x_t \mid x_v)
  = \prod_{i=1}^{L}
    p(x_{t,i} \mid x_v, x_{t,<i}),
\end{align}

where $x_{t,<i}$ denotes all text tokens preceding position $i$.

\subsection{Supervised fine-tuning}
During fine-tuning, the model continues training to predict each assistant token.
Given an image–question–answer triple $(v, q, a)$, the inputs are processed as follows:
the image is encoded into visual tokens $x_v = f_{\text{vision}}(v)$,
the question into text tokens $x_q = f_{\text{tokenize}}(q)$,
and the answer into target tokens $x_a = f_{\text{tokenize}}(a)$.

The model predicts each assistant token
$x_{a,i}$ conditioned on the visual embeddings, the question tokens,
and all previously generated assistant tokens:

\begin{align}
p(x_a \mid x_v, x_q)
  = \prod_{i=1}^{L_a}
    p(x_{a,i} \mid x_v, x_q, x_{a,<i}),
\end{align}

where $L_a$ is the number of assistant tokens.
The supervised fine-tuning objective minimizes the negative log-likelihood
of the reference assistant tokens:

\begin{align}
\mathcal{L}_{\text{SFT}}
  = - \sum_{i=1}^{L_a}
      \log p(x_{a,i} \mid x_v, x_q, x_{a,<i}).
\end{align}

\subsection{Group Relative Policy Optimization}

For a question-answer pair $(q,a)$, G responses $\{o_i\}_{i=1}^G$ are independently sampled from the old policy model $\pi_{\theta_{\text{old}}} $. Every response gets rewarded individually resulting in $\{R_i\}_{i=1}^G$. The advantage $\hat{A}_{i,t}$ is computed with respect to the group as described in \equationref{eq:grpo_advantage}. The GRPO objective is described in \equationref{eq:grpo_objective}.

\begin{equation}
\begin{aligned}
\mathcal{L}_{\text{GRPO}}(\theta) 
&= 
\mathbb{E}_{(q,a)\sim\mathcal{D},\, \{o_i\}_{i=1}^G \sim \pi_{\theta_{\text{old}}}(\cdot|q)} 
\\
&\quad \Bigg[
\frac{1}{G} \sum_{i=1}^G \frac{1}{|o_i|} \sum_{t=1}^{|o_i|}
\Big(
\min\big(
r_{i,t}(\theta)\, \hat{A}_{i,t},\,
\text{clip}(r_{i,t}(\theta), 1 - \epsilon, 1 + \epsilon)\, \hat{A}_{i,t}
\big)
- \beta\, D_{\text{KL}}(\pi_\theta \| \pi_{\text{ref}})
\Big)
\Bigg],
\label{eq:grpo_objective}
\end{aligned}
\end{equation}

where

\noindent
\begin{minipage}{0.48\linewidth}
\begin{align}
r_{i,t}(\theta) &=
\frac{
\pi_\theta(o_{i,t} \mid q, o_{i,<t})
}{
\pi_{\theta_{\text{old}}}(o_{i,t} \mid q, o_{i,<t})
},
\end{align}
\end{minipage}
\hfill
\begin{minipage}{0.48\linewidth}
\begin{align}
\hat{A}_{i,t} &=
\frac{
R_i - \text{mean}(\{R_i\}_{i=1}^G)
}{
\text{std}(\{R_i\}_{i=1}^G)
},
\label{eq:grpo_advantage}
\end{align}
\end{minipage}
\vspace{1.5em}

We follow previous work~\citep{yu2025dapo} and use asymmetric clipping.

\section{Evaluation Metrics Details}
\label{sec:evaluation-metrics-details}
\begin{itemize}
    \item \textbf{BERTScore}~\citep{zhang2020BERTScore}: computes token-level semantic similarity using contextual embeddings from a pretrained model (see \tableref{suptab:bertscore-setting}).
    \item \textbf{ROUGE-L}~\citep{lin2004rouge}: evaluates the longest common subsequence between generated and reference texts.
\end{itemize}

\begin{itemize}
    \item \textbf{RadGraph F1}~\citep{jain2021radgraph, delbrouck2024radgraph}: measures structural overlap between predicted and ground-truth entities and their relations. We employ the \textbf{RadGraph-XL}~\citep{delbrouck2024radgraph} model and report the partial reward variant of the metric.
    \item \textbf{CheXbert F1} We report the macro F1 over the 14 labels.
    \item \textbf{RadCliQ}~\citep{yu2023evaluating} is a composite metric which first standardizes and then linearly combines RadGraph F1, CheXbert vector similarity, BERTScore, and BLEU to align better with radiologists. We use RadGraph-XL~\citep{delbrouck2024radgraph} as our chosen RadGraph model.
    \item \textbf{GREEN} (Generative Radiology Report Evaluation and Error Notation)~\citep{ostmeier2024green}: an LLM-as-Judge metric that identifies clinically significant errors and matched findings in generated reports.
\end{itemize}

\clearpage
\section{Comparison between orignal and our updated RadVLM.}
\label{sec:comparison-to-original-radvlm}
We compare our update RadVLM to the original RadVLM in \tableref{tab:radvlm-comparison-report-generation} and \tableref{tab:radvlm-comparison-visual-grounding}. The updated RadVLM, based on Qwen3-VL, outpeforms the original RadVLM in all metrics except for the anatomy visual grounding.

\begin{table}[htbp]
\setlength\tabcolsep{2pt}

 \floatconts
 {tab:radvlm-comparison-report-generation}%
 {\caption{\textbf{Updated RadVLM comparison Report Generation.} We compare the original RadVLM~\cite{deperrois2025radvlm} to the updated RadVLM based on Qwen3-VL on report generation.}}%
 {
 \begin{tabular}{@{}p{0.3\linewidth}
 p{0.06\linewidth}
 p{0.08\linewidth}
 p{0.08\linewidth}
 p{0.08\linewidth}
 p{0.08\linewidth}
 p{0.08\linewidth}
 p{0.08\linewidth}
 p{0.08\linewidth}@{}}
 \toprule
\bfseries Model & \bfseries Size (B) &\bfseries R-L & \bfseries B-S & \bfseries CXb-micro & \bfseries CXb-macro & \bfseries RGF1 & \bfseries GRN & \bfseries RCQ$\downarrow$ \\
RadVLM (original) &7& 25.4 & 51.9 & 46.4 & 28.9 & 18.2 & 27.7 & 1.16 \\
RadVLM (ours)  &8& 26.0 & 53.3 & 49.0 & 30.5 & 19.0 & 29.1 & 1.14 \\
\bottomrule
\end{tabular}}
\end{table}

\begin{table}[htbp]
\floatconts
 {tab:radvlm-comparison-visual-grounding}%
 {\caption{\textbf{Updated RadVLM comparison Visual Grounding.}. We compare the original RadVLM~\cite{deperrois2025radvlm} to the updated RadVLM based on Qwen3-VL on visual grounding.}}%
{\begin{tabular}{p{0.35\linewidth}
 p{0.13\linewidth}
 p{0.13\linewidth}
 p{0.13\linewidth}}
\toprule
\bfseries Model & \bfseries Anatomy & \bfseries Abnorm.  & \bfseries Phrase$_{\text{MS}}$  \\ 
\midrule
RadVLM (original) & 85.8 & 34.6 & 81.8\\ 
\midrule
RadVLM (ours) & 82.1 & 44.2 & 84.6 \\ 
\bottomrule
\end{tabular}}%
\end{table}

\clearpage
\section{Hugging Face Links}
\label{appendix-sec:hf-links}
We link to the model weights in \tableref{suptab:baseline-models-hugging-face}.
\begin{table}[h]
\floatconts
  {suptab:baseline-models-hugging-face}%
  {\caption{\textbf{Model Links.} Huggingface links for the models we use.}}%
  {\begin{tabular}{p{0.35\linewidth}
 p{0.6\linewidth}}
  \toprule
  \bfseries Model  & \bfseries Hugging Face \\
  \midrule
  Qwen3-VL-8B-Instruct & \url{https://huggingface.co/Qwen/Qwen3-VL-8B-Instruct} \\
  Qwen3-VL-8B-Thinking & \url{https://huggingface.co/Qwen/Qwen3-VL-8B-Thinking} \\
  Qwen3-VL-235B-A22B-Instruct & \url{https://huggingface.co/Qwen/Qwen3-VL-235B-A22B-Instruct-FP8} \\
  CheXagent-2-3b & \url{https://huggingface.co/StanfordAIMI/CheXagent-2-3b}\\
  MedGemma-4b-pt & \url{https://huggingface.co/google/medgemma-4b-pt}\\
  MedGemma-27b-it & \url{https://huggingface.co/google/medgemma-27b-it} \\
  DeepMedix-R1& \url{https://huggingface.co/Qika/DeepMedix-R1} \\
  MAIRA-2& \url{https://huggingface.co/microsoft/maira-2} \\
  Llava-Rad & \url{https://huggingface.co/microsoft/llava-rad} \\
  \bottomrule
  \end{tabular}}
\end{table}

\clearpage
\section{verl parameters}
We list the parameters we use for verl in \tableref{suptab:verl-parameters}.

\begin{table}[htbp]
\floatconts
  {suptab:verl-parameters}%
  {\caption{\textbf{Parameters used with verl.}. If there are two values then the first one is for no thinking and the second one is for with thinking.}}%
  {\begin{tabular}{lll}
  \toprule
  \bfseries Group & \bfseries Name & \bfseries Value\\
  \midrule
  \multirow{3}{*}{Data}
 & batch\_size & 512 \\
 & max\_prompt\_length & 4096 \\
 & max\_response\_length & 1024 / 4096 \\ 
 & min\_pixel & 1024 \\
 & max\_pixel & 451584 \\
 \midrule

 \multirow{8}{*}{Actor}
 & lr & 1e-6 \\
 & ppo\_mini\_batch\_size & 128 \\
 & ppo\_micro\_batch\_size\_per\_gpu & 4 \\
 & use\_kl\_loss & True \\
 & kl\_loss\_coef & 0.01 \\
 & kl\_loss\_type & low\_var\_kl \\
 & clip\_ratio\_low & 0.20 \\
 & clip\_ratio\_high & 0.28 \\ \midrule

\multirow{3}{*}{Rollout}
 & log\_prob\_micro\_batch\_size\_per\_gpu & 16 \\
 & n: group size per prompt & 8 \\
 & log\_prob\_micro\_batch\_size\_per\_gpu (ref) & 16 \\\midrule

\multirow{4}{*}{Trainer}
 & n\_gpus\_per\_node & 4 \\
 & nnodes & 8 \\
 & save\_freq & 20 \\
 & test\_freq & 20 \\

\bottomrule
  \end{tabular}}
\end{table}

\clearpage
\section{BERTScore settings}
We list the BERTScore setting in \tableref{suptab:bertscore-setting}.

\begin{table}[htbp]
\floatconts
  {suptab:bertscore-setting}%
  {\caption{\textbf{Settings used with BERTScore.}}}%
  {\begin{tabular}{lll}
  \toprule
  \bfseries Name & \bfseries Value\\
  \midrule

  model\_type & distilbert-base-uncased \\
  num\_layers & 5  \\
  all\_layers & False \\
  idf & False \\
  lang & en \\
  rescale\_with\_baseline & True \\
\bottomrule
  \end{tabular}}
\end{table}

\clearpage
\section{Prompts for Models}
\label{appendix-sec:model-prompts}

We list the prompts we use for the various models. Summary of the input prompts (or templates / description of how the inputs are processed) employed for inference by each compared model. These templates were obtained from released material (paper, GitHub repo, etc.) when available; otherwise, RadVLM's prompts were used (for LLaVA-OV and LLaVA-Med). “HF” denotes HuggingFace. For all RadVLM variants and for the RL training of Qwen3-VL models we use the corresponding RadVLM prompts.
\subsection{Report Generation}
\begin{tcolorbox}[
  title=CheXagent-2,
  fonttitle=\bfseries,
  colframe=black,
  colback=white,
  toptitle=1mm,
  bottomtitle=1mm,
  top=2mm,
  breakable,
]
Write an example findings section for the CXR
\end{tcolorbox}

\begin{tcolorbox}[
  title=MAIRA-2,
  fonttitle=\bfseries,
  colframe=black,
  colback=white,
  toptitle=1mm,
  bottomtitle=1mm,
  top=2mm,
  breakable,
]
HF template: processed inputs passed directly to model (no explicit prompt)
\end{tcolorbox}

\begin{tcolorbox}[
  title=DeepMedix-R1,
  fonttitle=\bfseries,
  colframe=black,
  colback=white,
  toptitle=1mm,
  bottomtitle=1mm,
  top=2mm,
  breakable,
]
\textless{}image\textgreater{}~Please act as an experienced radiologist and generate the ``FINDINGS'' section of an X-ray report based on the provided image(s). Carefully examine the image(s) and describe all observed anatomical structures and abnormalities in a systematic and objective manner. You FIRST think about the reasoning process as an internal monologue and then provide the final answer. The reasoning process MUST BE enclosed within \textless{}think\textgreater{} \textless{}/think\textgreater{} tags. During this reasoning process, prioritize analyzing the local regions of the image by leveraging the bounding box coordinates in the format [x\textunderscore min, y\textunderscore min, x\textunderscore max, y\textunderscore max]. The final answer MUST BE put in \textbackslash{}boxed\{\}. An example is like: \textless{}think\textgreater{} reasoning process 1 with [x\textunderscore min1, y\textunderscore min1, x\textunderscore max1, y\textunderscore max1]; reasoning process 2 with [x\textunderscore min2, y\textunderscore min2, x\textunderscore max2, y\textunderscore max2] \textless{}/think\textgreater{}. The answer is: \textbackslash{}boxed\{answer\}.
\end{tcolorbox}

\begin{tcolorbox}[
  title=LLaVA-Rad,
  fonttitle=\bfseries,
  colframe=black,
  colback=white,
  toptitle=1mm,
  bottomtitle=1mm,
  top=2mm,
  breakable,
]
\textless{}image\textgreater{}\textbackslash nDescribe the findings of the chest x-ray.\textbackslash n
\end{tcolorbox}

\begin{tcolorbox}[
  title=google/medgemma-27b-it,
  fonttitle=\bfseries,
  colframe=black,
  colback=white,
  toptitle=1mm,
  bottomtitle=1mm,
  top=2mm,
  breakable,
]
\textless{}image\textgreater{} You are an expert radiologist. Please succinctly describe the findings for the above chest x-ray.
\end{tcolorbox}

\begin{tcolorbox}[
  title=google/medgemma-4b-pt,
  fonttitle=\bfseries,
  colframe=black,
  colback=white,
  toptitle=1mm,
  bottomtitle=1mm,
  top=2mm,
  breakable,
]
\textless{}image\textgreater{} findings:
\end{tcolorbox}

\begin{tcolorbox}[
  title=Qwen3-VL-8B-Instruct and Qwen3-VL-8B-Thinking,
  fonttitle=\bfseries,
  colframe=black,
  colback=white,
  toptitle=1mm,
  bottomtitle=1mm,
  top=2mm,
  breakable,
]
\textless{}image\textgreater\textbackslash nPlease write a radiology report for this Chext X-ray.\textbackslash n\textbackslash nIt should be one unstructured paragraph of findings only: concise, natural clinical language, objective, declarative sentences describing visible features only, suitable for a radiology findings report using standard radiology phrasing.
\end{tcolorbox}

\subsection{Visual Grounding}

\begin{tcolorbox}[
  title=MAIRA-2,
  fonttitle=\bfseries,
  colframe=black,
  colback=white,
  toptitle=1mm,
  bottomtitle=1mm,
  top=2mm,
  breakable,
]
HF template: processed inputs passed directly to model
\end{tcolorbox}

\begin{tcolorbox}[
  title=Qwen3-VL-8B-Instruct and Qwen3-VL-8B-Thinking,
  fonttitle=\bfseries,
  colframe=black,
  colback=white,
  toptitle=1mm,
  bottomtitle=1mm,
  top=2mm,
  breakable,
]
\textless{}image\textgreater\textbackslash nCan you tell me where [abnormality] is? Answer with bounding boxes only in the format [$x_{min}, y_{min}, x_{max}, y_{max}$], values normalized between 0 and 1 and rounded to two decimals. Multiple boxes should be separated by 'and'.
\end{tcolorbox}

\clearpage
\section{Qwen3-VL-235B Performance on Tasks}

We create an overview of the performance of Qwen3-VL-235B models listed in the Qwen3-VL technical report~\cite{bai2025qwen3vltechnicalreport} in \tableref{tab:qwen-q3vl-summary}.

\begin{table}[h]
\floatconts
  {tab:qwen-q3vl-summary}%
  {\caption{Summary of Qwen3-VL-235B-Thinkng and -Instruct results across benchmark groups. Scores are taken from the Qwen3-VL technical report~\cite{bai2025qwen3vltechnicalreport} and directly averaged directly within each group to provide a high level overview, despite differences in scales across benchmarks. Further, number of individual benchmarks that favor Instruct, Thinking, or Tie respectively are listed.}}%
  {\begin{tabular}{p{0.37\linewidth}p{0.1\linewidth}p{0.1\linewidth}p{0.1\linewidth}p{0.1\linewidth}p{0.05\linewidth}}
  \toprule
  \bfseries Benchmark & \bfseries Instruct & \bfseries Thinking & \bfseries Favor Instruct & \bfseries Favor Thinking & \bfseries Ties\\
  \midrule
  STEM Puzzle & 62.31 & \textbf{66.49} & 1& \textbf{13} & 0\\
  General VQA & \textbf{79.76} & 79.74 & \textbf{3} & 2 & 0\\
  Alignment & 54.33 & \textbf{55.97} & 0 & \textbf{2} & 1\\
  Document Understanding & \textbf{131.25} & 128.11 & \textbf{6} & 5 & 2\\
  2D/3D Grounding & \textbf{57.13} & 54.77 & \textbf{4} & 2 & 0\\
  Embodied/Spatial Understanding & 66.40 & \textbf{68.12} & 1 & \textbf{4} & 0 \\
  Multi-Image & 71.85 & \textbf{73.60} & \textbf{1} & \textbf{1} & 0 \\
  Video Understanding & 73.61 & \textbf{73.74} & \textbf{5} & 2 & 0\\
  Perception with Tool & \textbf{87.17} & 82.27 & \textbf{3} & 0 & 0\\
  Multi-Modal Coding & \textbf{80.77} & 79.20 & \textbf{2} & 1 & 0 \\
  Multi-Modal Agent & 50.58 & \textbf{52.46} & 2 & \textbf{3} & 0\\
  \midrule
  Average & 79.26 & \textbf{79.43} & 28 & \textbf{35} & 3 \\
  \bottomrule
  \end{tabular}}
\end{table}

\clearpage
\section{Cold-Start Prompt}
\label{appendix-sec:cold-start-prompts}
\subsection{Report Generation}
We provide the prompt to generate the cold-start dataset. The ``guide`` (GUIDE) is adapted based on sections 1.2 - 1.10 of \url{https://www.ncbi.nlm.nih.gov/books/NBK553874/} with tables converted into markdown. The guide and ground truth are substituted into the prompt in their respective placeholders. Note that some whitespace has been removed in this box.

\begin{tcolorbox}[
  title=Report Generation Cold-Start Prompt,
  fonttitle=\bfseries,
  colframe=black,
  colback=white,
  toptitle=1mm,
  bottomtitle=1mm,
  top=2mm,
  breakable,
]
You are an expert thoracic radiologist analyzing a chest radiograph.

Your task is to perform a **structured, detailed, and systematic reasoning process** based on the GUIDE below and the GROUND TRUTH report.

---

\#\#\# INSTRUCTIONS

1. **Follow the GUIDE carefully and comprehensively.**
   - Proceed systematically through every anatomic and technical category described in the GUIDE.
   - For each category, analyze findings in a detailed, clinically reasoned manner.

2. **Use of the Ground Truth**
   - Treat each *positive* finding explicitly stated in the Ground Truth as present with the same certainty level.
   - Treat each *negative* finding explicitly stated in the Ground Truth as absent with the same certainty level.
   - Treat each *uncertain* finding in the Ground Truth as indeterminate with the same certainty level.
   - You must explicitly mention each item in the Ground Truth in the appropriate section. Do not skip any item, even if it seems incidental.
   - **Crucially: if a finding is *not mentioned* in the Ground Truth, do *not* assume absence.** Silence in the Ground Truth means the status is unknown until you assess the image.
   - Base your reasoning primarily on the image. If the image evidence is insufficient, label the finding as **indeterminate** rather than inferring from Ground Truth silence.
   - The output must not mention or allude to having access to the Ground Truth.

3. **Certainty Language**
   - When appropriate, qualify statements as one of:
     - **Present (positive)** — clear imaging evidence supports the finding.
     - **Absent (negative)** — clear imaging evidence supports absence.
     - **Indeterminate/uncertain** — imaging is equivocal, limited, or non-diagnostic for that point.
     - **Not assessable** — technical factors preclude evaluation.
   - Choose the least speculative label when evidence is limited.

4. **Output Requirements**
   - Provide **only your structured reasoning process**, not a final radiology report.
   - The reasoning should be clear, logically structured, and clinically grounded.
   - Do not mention instructions, the GUIDE, or any external references.
   - Do not reference the ground truth in the structured reasoning process. Write as if directly interpreting the image.
   - Discuss the following in order, using paragraph form (not lists):
     - **Technical quality** (positioning, penetration, motion, lung volumes, artifacts)
     - **Support and monitoring devices and other foreign bodies/surgical materials**
     - **Chest wall**
     - **Mediastinum** (heart, great vessels, masses, lines/stripes/interfaces, calcification, pneumomediastinum)
     - **Hila**
     - **Lungs** (volumes, atelectasis, air space opacities, interstitial changes, nodules/masses, abnormal lucency)
     - **Airways**
     - **Pleura and diaphragm**

5. **Tone and Style**
   - Be explicit, methodical, and clinically precise.
   - Use full sentences and natural medical phrasing.
   - Demonstrate radiologic reasoning rather than just listing facts.

---

\#\#\# GUIDE

\{GUIDE\}

---

\#\#\# GROUND TRUTH

\{GROUND\_TRUTH\}

\end{tcolorbox}

\subsection{Visual Grounding}

We provide the prompt to generate the cold-start dataset for visual grounding. The relevant parts are substituted into the prompt in their respective placeholders. Note that some whitespace has been removed in this box.

\begin{tcolorbox}[
  title=Report Generation Cold-Start Prompt,
  fonttitle=\bfseries,
  colframe=black,
  colback=white,
  toptitle=1mm,
  bottomtitle=1mm,
  top=2mm,
  breakable,
]
\#\#\# System
You are an expert chest-radiology vision–language model specialized in **visual grounding**.  
Your main goal is to produce an extensive, detailed reasoning trace that describes **how you locate the visual evidence** in the chest X-ray that supports the question, before revealing the answer.

For every case you will receive  
- one chest X-ray image  
- a clinical question about that image  
- the ground-truth answer to that question  

**Your task**

1. Carefully examine the image and reason **step-by-step** about the **spatial evidence** related to the question.  
   - Focus on **where** the relevant anatomical structure or abnormality is located.  
   - Describe **which region(s)** of the image you examine, how you navigate the X-ray, and how you narrow down to specific areas.  
   - Explain spatial relationships (left/right, upper/mid/lower zones, central/peripheral, anterior/posterior) and link findings to nearby anatomical landmarks (diaphragm, heart border, ribs, hilum, costophrenic angles, etc.).  
   - Discuss the **bounding-box level reasoning**: which portion of the image would contain the key feature.  
   - Mention what visual patterns guide your search (opacity, lucency, silhouette loss, contour shape, asymmetry).  
   - Consider alternative locations or causes and justify why you include or exclude them.  
   - Be **very detailed and verbose** — the reasoning should read like a radiologist’s internal monologue mapping the image to bounding boxes.  
   - You must complete this reasoning **fully** before revealing the answer.

2. Write this full reasoning inside one continuous block that ends with **`\textless{}/think\textgreater{}`**.  
   - Do **not** output the answer or hint at it before this token.  
   - The reasoning block should be long, coherent, and spatially rich.

3. Immediately after `\textless{}/think\textgreater{}`, output the **gold answer exactly as given** in the input (copy it verbatim).  
   - Do not add any commentary or text outside these two blocks.  

---

\#\#\# User

Question: \{QUESTION\}

Gold Answer (copy verbatim after the thinking block):  
\{ANSWER\_GT\}

---

\#\#\# Assistant
\textless{}!-- your richly detailed spatial reasoning trace describing how you identify the relevant region(s) --\textgreater{}
\textless{}/think\textgreater{}
\{ANSWER\_GT\}

\end{tcolorbox}

\clearpage
\section{Distilling Cold-Start from Qwen3-VL-32B-Thinking}
\label{sec-app:cold-start-distilling}
We also tried distilling from Qwen3-VL-32B-Thinking. While these results are preliminary, we think they might be helpful for future work.

\subsection{Report Generation}
In addition to the cold-start generation method where we provide the ground truth, we also tried to distill from Qwen3-VL-32B-Thinking, by sampling 8 times per image (the ground truth is not provided) and then selecting the best answer (with corresponding thinking) according to RadCliQ. First, we tried a ``One-Stage'' approach, where we directly ask the model to output the findings in a format that is similar to the findings in our dataset. This appeared to bias the model, so we tried another approach. Second, we tried a ``Two-Stage'' approach. We visualize the rewards in \figureref{fig:cold-start-distillation-report-generation} and observe that the approach that is provided the guide and ground truth (``RadVLM-Thinking'') performs has higher rewards than the ``Two-Stage'' approach, which in return has higher rewards than the ``One-Stage'' approach. Please keep in mind, that a small percentage of the Thinking outputs get assigned a score of $-3$ because they do not produce a final answer and this. The evolution of the response length is shown in \figureref{fig:cold-start-distillation-report-generation-length}. We describe the ``One-Stage'' and``Two-Stage'' approaches next.
\begin{figure}[htbp]
\floatconts
  {fig:cold-start-distillation-report-generation}
   {\caption{ \textbf{Comparison of Cold-Starts: RadCliQ reward.}   
    }}  {\includegraphics[width=0.9\linewidth]{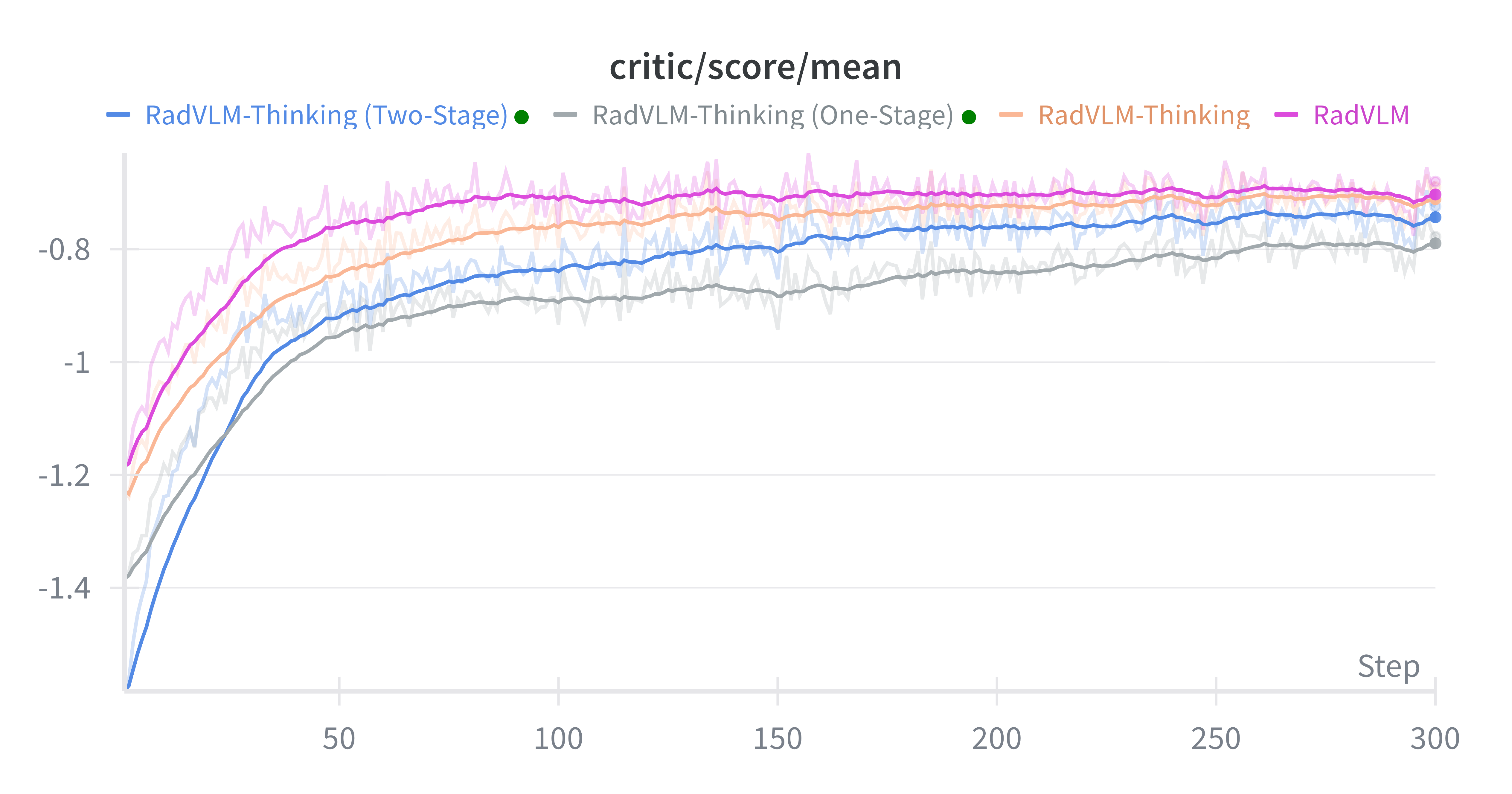}}
\end{figure}

\begin{figure}[htbp]
\floatconts
  {fig:cold-start-distillation-report-generation-length}
  {\caption{ \textbf{Comparison of Cold-Starts: evolution of response Length.}  
    }}  {\includegraphics[width=0.9\linewidth]{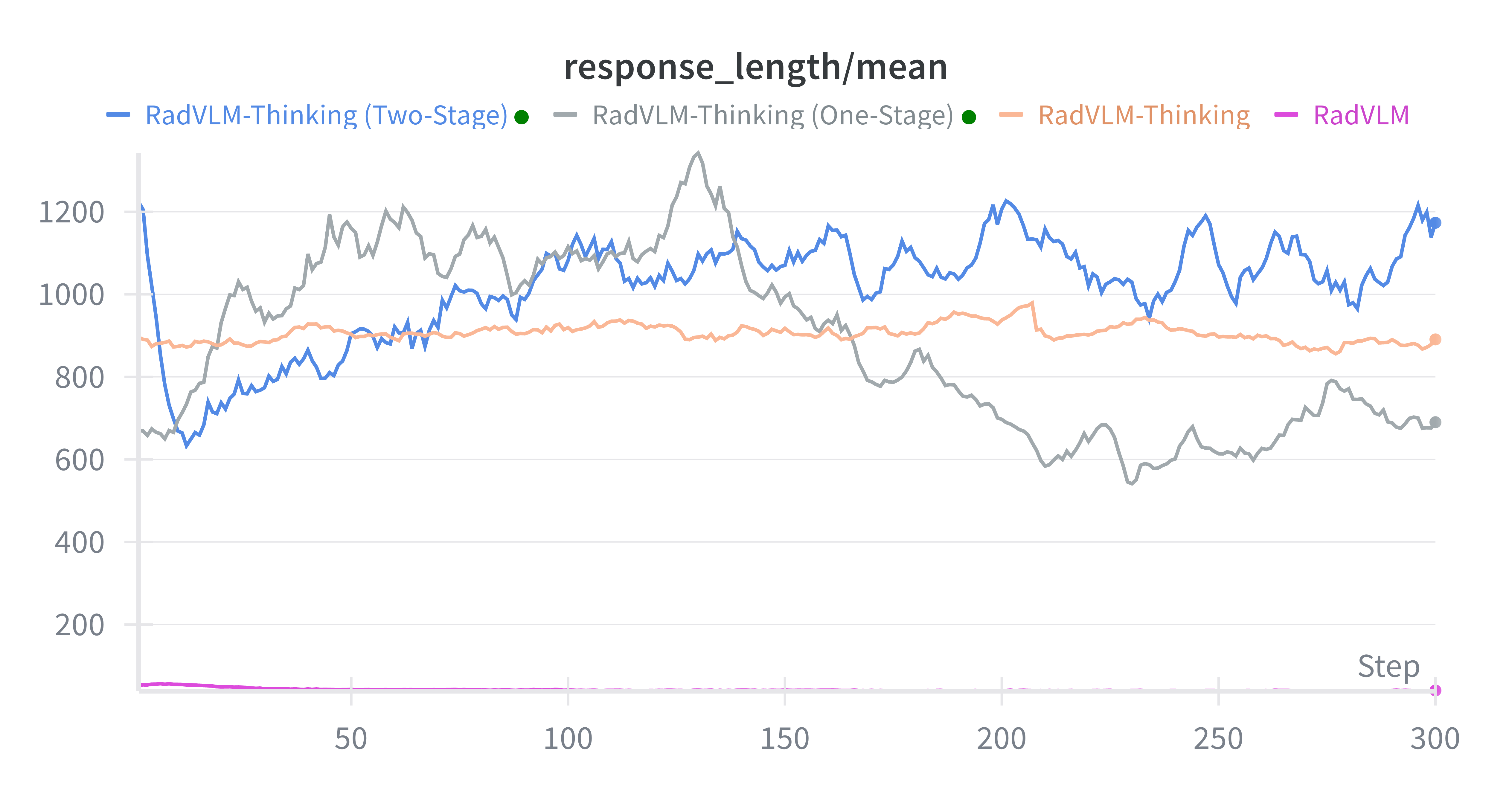}}
\end{figure}

\subsubsection{One-Stage}

For image $i$, we generate 8 responses from Qwen3-VL-32B-Thinking with the RadVLM prompts, resulting in thinking$_{i, j}$ and an answer$_{i, j}$. For image $i$ we then select the best answer$_{i, j}$ over j, according to RadCliQ, resulting in the best answer$_{i}$, and the corresponding thinking$_{i}$. We then use (thinking$_{i}$, answer$_{i}$) as a ``One-Stage'' cold-start datapoint.

\begin{tcolorbox}[
  title=One-Stage Prompt,
  fonttitle=\bfseries,
  colframe=black,
  colback=white,
  toptitle=1mm,
  bottomtitle=1mm,
  top=2mm,
  breakable,
]
You are an expert thoracic radiologist analyzing a chest radiograph.
\\
\\
After finishing your reasoning, close it with \textless{}/think\textgreater{} and then produce the final answer:\\
- One unstructured paragraph of Findings only: concise, natural clinical language, objective, declarative sentences describing visible features only, suitable for a radiology findings report using standard radiology phrasing.\\
- Include positives and relevant negatives that are visually supported.\\
- Use concise, compact, natural phrasing consistent with real radiology reports, keep Findings descriptive (no differential diagnoses, management, no overall judgment).\\
- Do not include headers, bullets, impressions, recommendations, or any mention of your thinking process.
\end{tcolorbox}

\subsubsection{Two-Stage}

For image $i$, we generate 8 responses from Qwen3-VL-32B-Thinking with the RadVLM prompts, resulting in thinking$_{i, j}$ and an answer$_{i, j, 1}$. The answer might look totally different than what we would expect, so for every response we let Qwen3-VL-32B-Instruct rewrite answer$_{i, j, 1}$ into answer$_{i, j, 2}$ with the following Two-Stage Rewriting Prompt. For image $i$ we then select the best answer$_{i, j, 2}$ over j, according to RadCliQ, resulting in the best answer$_{i}$, and the corresponding thinking$_{i}$. We then use (thinking$_{i}$, answer$_{i}$) as a ``Two-Stage`` cold-start datapoint.

\begin{tcolorbox}[
  title=Two-Stage Rewriting Prompt,
  fonttitle=\bfseries,
  colframe=black,
  colback=white,
  toptitle=1mm,
  bottomtitle=1mm,
  top=2mm,
  breakable,
]
You are a radiology expert.\\
\\
You are given a full radiology report. Your task is to generate the Findings only.\\
\\
Requirements for the Findings:\\
- Output exactly one unstructured paragraph of Findings text only.\\
- Do not include any label such as "Findings:", no headers, no section titles, no bullets, no tables.\\
- Do not include explanations.\\
- Use concise, natural clinical language with objective, declarative sentences.\\
- Describe only visible imaging features. Do not include differential diagnoses, do not include disease labels that are not explicitly present in the original report, do not include management, do not include overall impressions.\\
- Lead with devices and any urgent or acute abnormalities when present.\\
- Include salient positive findings and relevant negatives that are documented in the original report.\\
- Use standard radiology phrasing consistent with real radiology reports.\\
- Write the Findings in the same language as the input report.\\
- Do not add information that is not supported by the original report. Do not hallucinate findings.\\
\\
Radiology report:\\
``````\\
\{answer$_{i, j, 1}$\}\\
``````\\
\end{tcolorbox}

\end{document}